\documentclass[11pt]{article}

\usepackage[final]{acl}

\usepackage{times}
\usepackage{latexsym}

\usepackage[T1]{fontenc}

\usepackage[utf8]{inputenc}

\usepackage{microtype}

\usepackage{inconsolata}

\usepackage{amsmath}
\usepackage{amssymb}
\usepackage{booktabs}
\usepackage{float}
\usepackage{graphicx}
\usepackage[none]{hyphenat}
\title{Forgotten Words: Benchmarking NeoBERT for Dementia Detection in Low-Resource Conversational Filipino and English Speech}

\author{
\textbf{Rez Samantha Z. Floresca$^{1}$}\thanks{Lead author.},
\textbf{Edric Castel C. Hao$^{2}$}\thanks{Provided research mentorship and guidance.}\thanks{This work does not relate to the author's position at Analog Devices, Inc.},
\textbf{Hannah Grachiella Buñales$^{1}$}, \\
\textbf{Chelsea Dominique E. Temprosa$^{1}$},
\textbf{Georgianna Z. Reyes$^{1}$},
\textbf{Kervin Gabriel L. Chua$^{1}$} \\
$^{1}$Ateneo de Manila Senior High School \quad $^{2}$Analog Devices, Inc. \\
\texttt{rez.samantha.floresca@student.ateneo.edu}
}

\usepackage{hyperref}
\begin{document}
\maketitle
\begin{abstract}
Dementia detection from spontaneous speech offers a scalable approach to cognitive screening, yet NLP systems remain predominantly English-centric. This limitation is especially acute in the Philippines, where Filipino–English code-switching is pervasive and no prior work has addressed NLP-based dementia detection. We present the first systematic evaluation of transformer-based dementia detection in Filipino speech and the first assessment of NeoBERT in a clinical NLP setting. To separate language from domain effects, we construct a parallel bilingual dataset of 4,000 DementiaBank-derived transcripts, with Filipino translations produced manually to preserve discourse-level markers of cognitive decline. We evaluate five model families, TF-IDF + LogReg, BERT, NeoBERT, XLM-R, and RoBERTa-Tagalog, under monolingual, zero-shot cross-lingual, and bilingual fine-tuning settings. We find that in-domain performance does not transfer across languages, with English-trained BERT dropping to Macro-F1 = 0.455 on Filipino, and that architectural modernization alone does not improve robustness. Bilingual fine-tuning, however, eliminates cross-lingual degradation across all transformer models, converging to Macro-F1 = 0.969--0.973. These results suggest that multilingual clinical NLP performance is driven primarily by linguistic coverage during training rather than model scale or architecture.
\end{abstract}
\section{Introduction}

Dementia detection from speech is a promising direction for scalable cognitive screening, as linguistic and discourse-level patterns often reflect early cognitive decline \cite{Peled-Cohen:25}. However, most NLP systems for this task are developed and evaluated primarily on English-language data, limiting their robustness in multilingual and low-resource settings. This limitation is particularly relevant in the Philippines, where everyday speech frequently involves Filipino–English code-switching and informal conversational structure. As a result, it remains unclear whether failures arise from cross-lingual transfer limitations or English-centric biases in model pretraining, motivating our focus on conversational transcripts to isolate discourse-level markers of cognitive decline.

Progress in Filipino NLP, while growing, has been largely confined to classification tasks on news articles and social media posts, including sentiment analysis \cite{Villavicencio2021Twitter}, named entity recognition \cite{miranda2023tlunified}, and hate speech detection \cite{cruz-cheng-2022-improving}. Even recent evaluation benchmarks such as Batayan \cite{batayan2025} and FilBench \cite{filbench2025}, while broader in scope, remain anchored to common written-text tasks such as reading comprehension, natural language inference, and generation. Critically, none of this work addresses naturalistic speech, clinical discourse, or cognitive diagnostic tasks. In particular, Tagalog–English code-switching, pervasive in informal and clinical Filipino communication \cite{herrera2022tweettaglish}, is systematically absent from existing evaluations. As a result, it remains unknown whether transformer-based models can capture the discourse-level degradation characteristic of dementia in Filipino speech.

Additionally, transformer-based encoders have become the dominant paradigm in clinical NLP \cite{devlin2019BERT, pappagari2020speaker}, yet their application to dementia detection remains largely confined to architectural variants that differ only in pretraining data. ClinicalBERT \cite{alsentzer2019clinical}, AD-BERT \cite{mao2023adbert}, and mBERT \cite{pires2019multilingual} all inherit the original BERT design unchanged. This leaves open the question of how architectural modifications affect the extent to which models capture cognitively diagnostic linguistic properties relevant to dementia detection.

In this work, we study dementia detection under controlled bilingual conditions using conversational speech transcripts in English and Filipino. The full experimental setup is described in Section~3.

Our contributions are threefold: (1)  We present the first evaluation of NLP-based dementia detection in Filipino speech, using conversational transcripts manually translated from English under controlled domain conditions. (2) We provide the first empirical characterization of NeoBERT's behavior under clinical and cross-lingual conditions.  (3) We analyze cross-lingual transfer from English to Filipino under controlled conditions, benchmarking English-only models against multilingual (XLM-RoBERTa) and language-matched (RoBERTa-Tagalog) baselines to quantify the effect of English-centric pretraining under language shift. Our code is publicly available.\footnote{\url{https://github.com/rezsam09/Filipino-English-Dementia-Classification}}

\section{Related Work}

\subsection{Linguistic Markers of Cognitive Decline}

Spontaneous speech in dementia and mild cognitive impairment (MCI) exhibits reduced lexical diversity, increased disfluency, syntactic simplification, and degraded referential coherence, patterns that manifest measurably before clinical diagnosis \cite{Fraser:2016, Segkouli:25} and have been linked to early tau accumulation in language-relevant cortical regions \cite{Marier:26}. A recent systematic review confirms that these markers are consistently recoverable from text alone, with linguistic-only models reaching an average diagnostic accuracy of 83\% across 51 NLP studies of dementia detection \cite{Shankar:25}.

However, existing work is heavily concentrated on monolingual English speech elicited under controlled clinical tasks, most commonly the Cookie Theft picture description task from DementiaBank \cite{becker1994natural}. Bilingualism introduces competing forces: lifelong code-switching is associated with delayed dementia onset through enhanced cognitive reserve \cite{Berkes:22}, while the syntactic mixing and lexical borrowing characteristic of Taglish may alter the observable linguistic signals on which current models rely \cite{Flores:19}. This makes Filipino a theoretically significant yet entirely absent deployment context in the dementia NLP literature.

\subsection{Transformer-Based Models in Clinical NLP}

In dementia detection, most systems adapt BERT through domain-specific pretraining: ClinicalBERT \cite{alsentzer2019clinical} adapts to clinical notes, AD-BERT \cite{mao2023adbert} targets Alzheimer's progression prediction, and mBERT \cite{pires2019multilingual} extends to multiple languages. These models differ mainly in pretraining data while sharing the same architecture, leaving open the question of whether architectural changes improve cross-lingual clinical performance.

Representation pooling is another underexplored yet consequential design choice. Most existing systems default to \texttt{[CLS]}-token pooling, a choice optimized for the pretraining objective but not for aggregating task-relevant information across the full sequence. Reimers and Gurevych \shortcite{reimers2019sentence} demonstrate that mean pooling over final hidden states produces more stable and transferable sentence representations, particularly under domain and language shift, motivating our adoption of mean pooling across all transformer-based models.

\subsection{Cross-Lingual Dementia Detection}

Cross-lingual transfer in dementia detection has been explored almost exclusively between typologically similar European languages. The ADReSS-M challenge \cite{luz2024adressm}, including work by Tamm et al. \cite{tamm2023crosslingual} and \cite{deloach2026dementia}, established English-to-Greek transfer as a benchmark setting. Greek, however, shares significant Indo-European typological properties with English, limiting the generalizability of these findings to more distant transfer settings such as English-to-Filipino. Studies on dementia detection in low-resource, non-European languages remain rare. Adhikari et al. \shortcite{adhikari2021exploiting} applied NLP-based detection to English DementiaBank transcripts translated into Nepali, while Lu and Chen \shortcite{lu2025amis} extended this line of work to Amis, an indigenous Taiwanese language, finding that augmenting a small dataset of 80 scripts with translated English transcripts improved both robustness and accuracy. Neither study examines cross-lingual transfer nor evaluates multilingual pretraining strategies. No work has examined dementia detection in a Southeast Asian language or in any language spoken by a population where bilingualism and code-switching are the norm rather than the exception. Filipino, spoken by over 100 million people and characterized by pervasive Filipino–English code-switching \cite{herrera2022tweettaglish}, represents exactly this gap.

\subsection{Multilingual and Language-Matched Pretraining}

Multilingual pretraining has emerged as a common strategy for cross-lingual transfer in low-resource settings. XLM-RoBERTa \cite{conneau2020xlmr}, trained on over 2.5 TB of filtered CommonCrawl data across 100 languages, substantially outperforms mBERT on low-resource cross-lingual benchmarks. However, even strong multilingual models do not fully capture the lexical and syntactic properties of Filipino: RoBERTa-Tagalog \cite{cruz-cheng-2022-improving}, pretrained on TLUnified, a large-scale, thematically diverse Filipino corpus, demonstrates consistent gains over XLM-RoBERTa on Filipino classification benchmarks, suggesting that language-matched pretraining captures aspects of Filipino that multilingual models do not adequately model. Whether this advantage transfers to a clinical domain, where training data is scarce and domain shift compounds language shift, remains entirely unknown.

\subsection{Architectural Modernization of Encoder Models}

NeoBERT \cite{lebreton2025neobert} represents the most recent advance in encoder architecture, replacing absolute positional embeddings with Rotary Positional Embeddings (RoPE) \cite{su2021roformer}, adopting Pre-LayerNorm \cite{xiong2020layer} and RMSNorm \cite{zhang2019rmsnorm} for training stability, introducing SwiGLU in place of GELU activation \cite{shazeer2020glu}, and optimizing depth-to-width ratio for parameter efficiency. Importantly, it remains exclusively English-pretrained on RefinedWeb, a corpus of 600B tokens, 18 times larger than RoBERTa's training data. Despite outperforming BERT-large, RoBERTa-large, and ModernBERT on MTEB under identical fine-tuning conditions \cite{lebreton2025neobert}, NeoBERT has not been evaluated in any clinical or cross-lingual setting, leaving open the question of whether its architectural advances translate into robustness under the combined domain and language shift characteristic of low-resource clinical NLP.
\section{Experimental Design}
\begin{figure}[ht]
    \centering
    \includegraphics[width=1\linewidth]{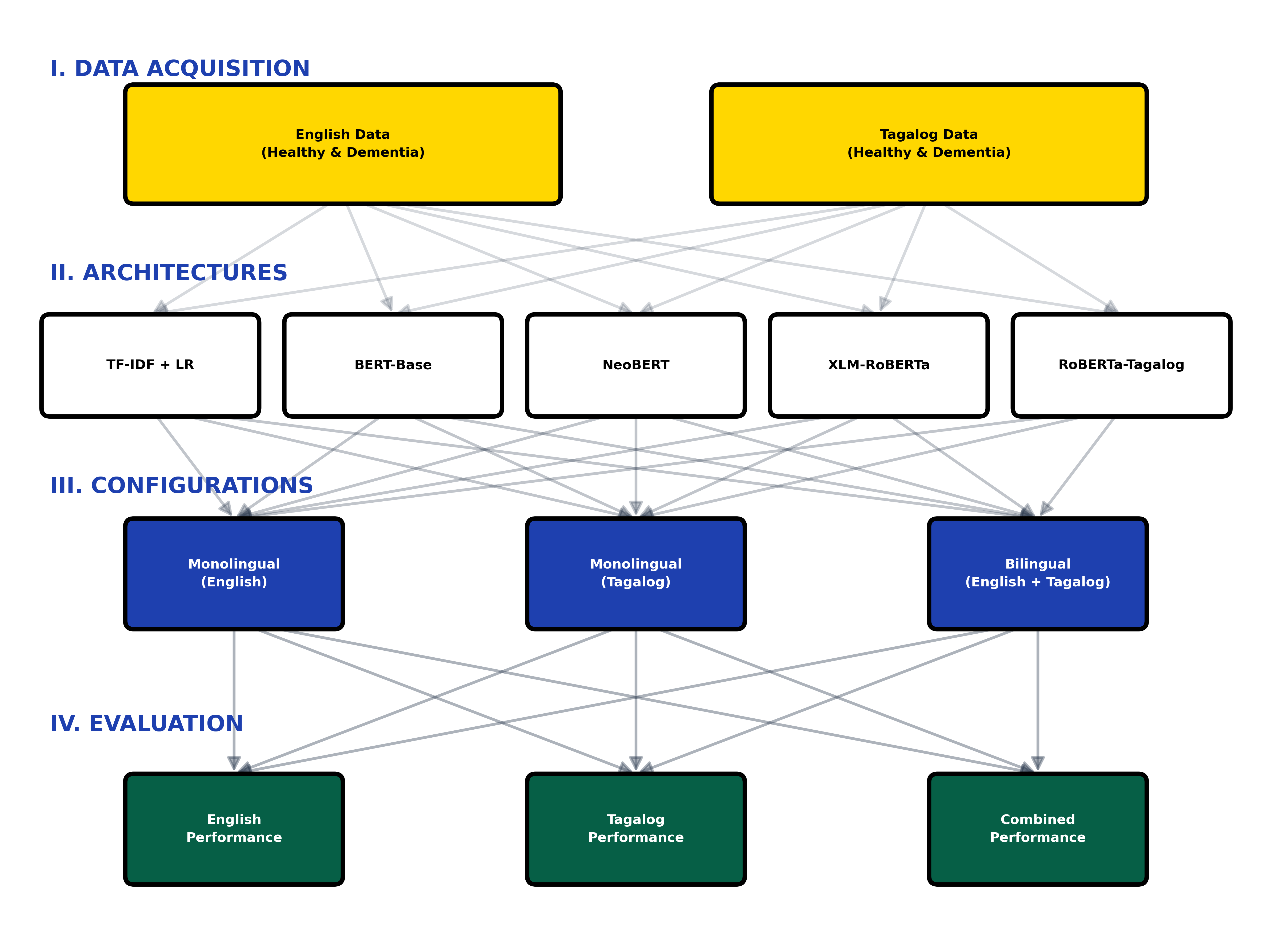}
    \caption{Overview of the experimental pipeline across datasets, model families, training configurations, and evaluation settings.}
    \label{fig:placeholder}
\end{figure}
Performance improvements in NLP are often confounded by simultaneous changes in data, preprocessing, and model architecture \cite{Sculley:2015}, a problem amplified in clinical NLP by small dataset sizes, domain sensitivity, and cross-linguistic variability \cite{Fraser:2016, Eyigoz:2020}. To mitigate this, all experiments use identical preprocessing pipelines with no external feature engineering beyond model-inherent representations, ensuring that observed performance differences reflect learned representations rather than preprocessing choices. The following subsections describe the dataset construction, preprocessing pipeline, model selection, and evaluation protocol.
\subsection{Dataset Construction and Preprocessing}
We construct a balanced bilingual dataset of 4,000 conversational transcripts for binary dementia classification, comprising 1,000 samples per class per language. All samples originate exclusively from DementiaBank \cite{becker1994natural}, ensuring that any observed cross-lingual performance degradation reflects linguistic shift rather than differences in topic distribution, discourse register, or clinical task structure.
\subsubsection{English Data Sources}
English dementia-positive and healthy control samples are drawn directly from DementiaBank \cite{becker1994natural}, a widely used corpus in clinical speech research consisting of elicited speech from structured clinical tasks, most notably the Cookie Theft picture description task. This controlled elicitation procedure yields comparable discourse conditions across participants \cite{Fraser:2016}, ensuring that class-level differences reflect genuine cognitive and linguistic variation rather than task-induced confounds.
\subsubsection{Filipino Data Sources}
Filipino samples are constructed through controlled manual translation of the complete set of 2,000 DementiaBank transcripts into Filipino, matching the English subset on clinical content, discourse structure, elicitation task, and class distribution. This ensures that cross-lingual evaluation is not confounded by asymmetric domain exposure. Manual translation is both necessary and principled given the absence of large-scale annotated clinical dementia corpora in Filipino \cite{Mave:2018}. Translators are given explicit instructions to retain discourse-level markers of cognitive decline, including repetitions, hesitations, false starts, and syntactic degradation. Machine translation is avoided, since neural systems are known to normalize disfluent speech toward fluent output \cite{salesky2019fluent}, systematically erasing the features that distinguish dementia from healthy speech. All samples are anonymized prior to training in accordance with standard clinical data handling protocols.
\subsubsection{Preprocessing Pipeline}
All transcripts undergo a standardized preprocessing pipeline applying Unicode normalization, whitespace normalization, and lowercasing. Disfluencies, including filled pauses, repetitions, and hesitation markers, are retained throughout, as these are established linguistic correlates of cognitive impairment \cite{Fraser:2016} and primary classification signals in dementia detection. Stemming, lemmatization, and syntactic parsing are avoided, since such transformations risk normalizing disfluency structures and degrading the diagnostic signal available to downstream models. All input sequences are truncated to a maximum length of 128 tokens, consistent with the short, task-elicited structure of DementiaBank transcripts and with findings that neurocognitive markers are recoverable from short windows of speech \cite{Fraser:2016}.
\subsubsection{Models and Baselines}
We evaluate five models across three pretraining paradigms. TF-IDF with Logistic Regression serves as a lexical baseline to assess whether surface-level token statistics suffice. BERT-base-uncased \cite{devlin2019BERT} and NeoBERT \cite{lebreton2025neobert} represent English-only pretraining, differing in architecture and corpus scale to isolate design capacity from language coverage. XLM-RoBERTa \cite{conneau2020xlmr} provides a 100-language multilingual baseline, while RoBERTa-Tagalog \cite{cruz-cheng-2022-improving} evaluates language-matched pretraining on native Filipino text. These models decompose cross-lingual robustness along two orthogonal axes: architecture and language coverage. Because all models share an identical pipeline, cross-lingual performance variations reflect pretraining language exposure rather than fine-tuning differences.

\noindent\textbf{Training Configurations:} Models are trained separately for three setups: (1) English-only ($\mathcal{D}_{\text{EN}}$), (2) Filipino-only ($\mathcal{D}_{\text{TL}}$), and (3) Bilingual ($\mathcal{D}_{\text{EN}} \cup \mathcal{D}_{\text{TL}}$), enabling controlled analysis of monolingual capacity, cross-lingual transfer, and bilingual transfer.
{\subsection{TF-IDF with Logistic Regression}}
\begin{figure}[H]
    \centering
    \includegraphics[width=1\linewidth]{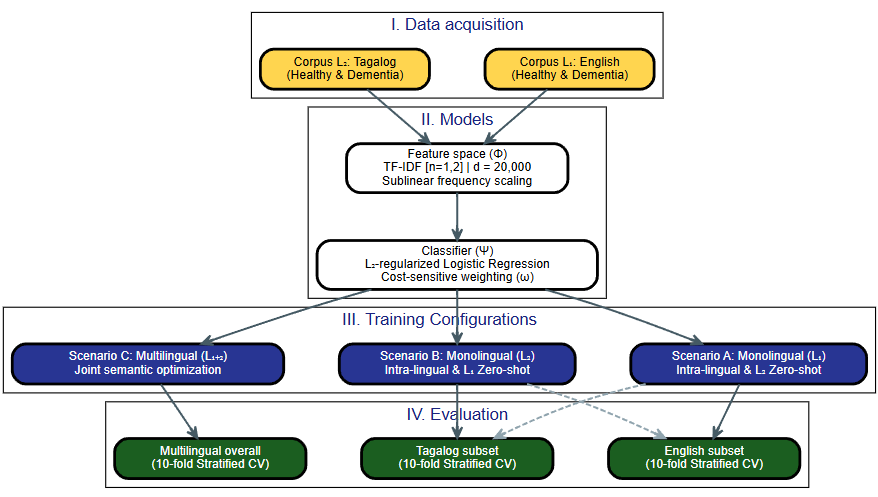}
    \caption{TF-IDF + Logistic Regression baseline training pipeline for evaluating lexical separability under cross-lingual and clinical domain shift conditions}
    \label{fig:placeholder}
\end{figure}
We use TF-IDF with Logistic Regression as a lexical baseline to assess whether dementia-related signals are captured by surface-level token statistics alone. TF-IDF produces sparse lexical representations by weighting tokens based on frequency statistics while ignoring word order and syntactic structure. Logistic Regression then learns a linear decision boundary over this high-dimensional feature space. This combination is widely used as an interpretable baseline in text classification tasks \cite{guo2020benchmarking, wang2012baselines, zhang2015character}. Performance of this model provides a reference for lexical separability: strong results suggest that class information is recoverable from surface-level features alone, whereas a weaker performance indicates the need for contextual modeling.
\subsubsection{TF-IDF Representation}
We construct unigram and bigram TF-IDF features ($n \in {1,2}$), capturing both lexical identity and short disfluency-related phrases. Bigrams are sufficient given the short, utterance-level structure of clinical transcripts and have been shown to yield consistent gains in text classification tasks \cite{wang2012baselines}.  We apply standard TF-IDF preprocessing settings: (1) sublinear term-frequency scaling ($1 + \log(tf)$) to reduce the influence of repeated tokens common in disfluent speech; (2) a minimum document frequency of 2 (absolute count) to remove rare transcription noise; (3) a maximum document frequency of 0.95 to filter non-informative high-frequency tokens; and (4) a maximum vocabulary size of 20,000, consistent with standard practice in text classification \cite{zhang2015character}. These constraints ensure a stable lexical feature space that is comparable across languages and avoids overfitting in low-resource settings.
\subsubsection{Logistic Regression Classifier}
We use $\ell_2$-regularized logistic regression due to its convex optimization properties and strong performance in high-dimensional sparse settings. Optimization is performed using the \texttt{liblinear} solver for efficient and deterministic convergence on sparse data, with a maximum of 2,000 iterations to ensure convergence.
\subsection{Transformer-Based Models}
\begin{figure}[ht]
    \centering
    \includegraphics[width=1\linewidth]{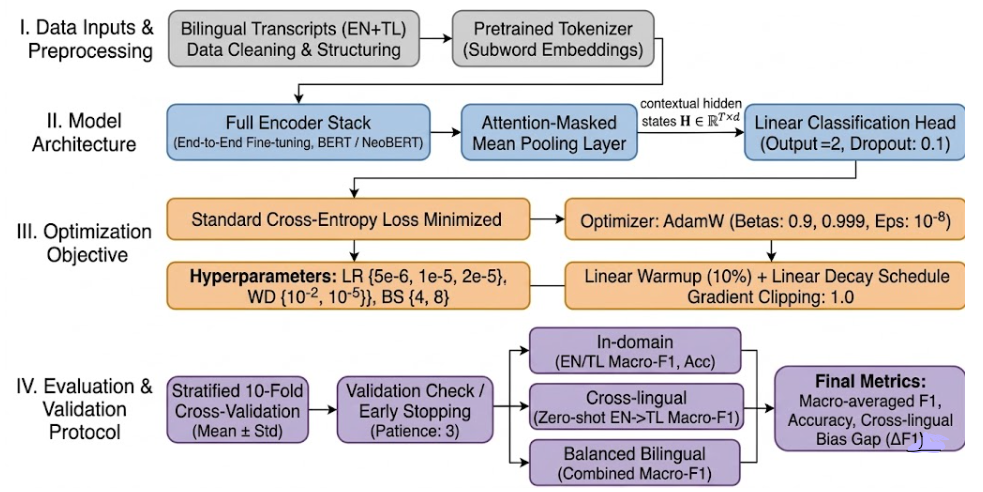}
    \caption{Transformer Models Fine-tuning Pipeline}
    \label{fig:placeholder}
\end{figure}

We fine-tune four transformer-based models for binary dementia classification: BERT-base-uncased, NeoBERT, XLM-RoBERTa, and RoBERTa-Tagalog. Together, these models span three pretraining paradigms: English-only, massively multilingual, and language-matched, enabling a systematic decomposition of cross-lingual robustness along both architectural and pretraining dimensions.
\subsubsection{Model Architecture and Fine-Tuning} 
All models follow a standard transformer-based classification pipeline. Given an input transcript $x$, tokens are first mapped to subword embeddings using a model-specific pretrained tokenizer, producing input IDs and attention masks. The resulting sequence is passed through the encoder, which is fine-tuned end-to-end without frozen layers.
A single sentence representation is obtained via attention-masked mean pooling over the final hidden states:
\[
h = \frac{\sum_{i=1}^{T} m_i H_i}{\sum_{i=1}^{T} m_i}
\]
\noindent
where $m_i$ denotes the attention mask, ensuring that padding tokens do not contribute to the sentence representation. Unlike CLS pooling, which is optimized for pretraining objectives, masked mean pooling aggregates distributed token-level information and has been shown to improve robustness under domain shift \cite{reimers2019sentence}.
The resulting representation is passed to a dropout layer ($p = 0.1$) followed by a linear classification head. The encoder is initialized from pretrained weights, while the classification head is randomly initialized.
\subsubsection{Optimization Objective}
We minimize standard cross-entropy loss:
\[
\mathcal{L}(\theta) = - \mathbb{E}_{(x,y)} \log p_{\theta}(y \mid x)
\]
Optimization is performed using AdamW \cite{loshchilov2019decoupled}, which decouples weight decay from gradient updates and improves generalization in transformer fine-tuning settings.
\subsubsection{Hyperparameter Selection}
Hyperparameters are selected via a systematic grid search designed to avoid loss divergence and unstable gradient updates on small clinical datasets.
The search space for all BERT models includes: (1) learning rates $\{5\times10^{-6}, 6\times10^{-6},1\times10^{-5}, 2\times10^{-5}, 3\times10^{-5}\}$
(2) weight decay $\{10^{-2}, 10^{-5}\}$
(3) batch sizes $\{4, 8\}$. These ranges are motivated by the constraints of low-resource clinical fine-tuning. Lower learning rates help prevent catastrophic forgetting of pretrained representations, while varying weight decay controls the balance between task-specific adaptation and robustness to noise introduced by manual transcription and translation. We use a stratified 70–15–15 train/validation/test split. The training set is used for model fitting, the validation set for early stopping, and the test set is used for hyperparameter selection.
Small batch sizes (${4, 8}$) are chosen to improve generalization through gradient stochasticity \cite{masters2018revisiting} while maintaining memory efficiency for fine-tuning NeoBERT. This also increases the number of parameter updates per epoch, which is beneficial in low-data regimes.

Table~\ref{tab:best-hparams-main} reports the optimal hyperparameter configurations identified via grid search for each architecture and training regime. Across models, learning rates in the range of $1 \times 10^{-5}$ to $3 \times 10^{-5}$ consistently yield stable fine-tuning, with lower rates ($1$--$2 \times 10^{-5}$) favored for architecturally complex or multilingual models such as XLM-RoBERTa and NeoBERT, likely reflecting greater sensitivity to catastrophic forgetting. NeoBERT consistently selects a batch size of 4 across all regimes, suggesting that higher gradient stochasticity benefits its fine-tuning under low-resource conditions. Weight decay varies across configurations, with stronger regularization ($1 \times 10^{-2}$) generally preferred in monolingual English and Filipino settings, while smaller values ($1 \times 10^{-5}$) are selected more frequently in bilingual and cross-lingual regimes, consistent with the need for greater representational flexibility under mixed-language supervision.
\begin{table}[ht]
\centering
\small
\resizebox{\columnwidth}{!}{%
\begin{tabular}{llccc}
\hline \hline
\textbf{Architecture} & \textbf{Regime} & \textbf{Batch} & \textbf{LR} & \textbf{WD} \\
\hline
BERT & English & 8 & $2 \times 10^{-5}$ & $1 \times 10^{-2}$ \\
BERT & Tagalog & 4 & $3 \times 10^{-5}$ & $1 \times 10^{-5}$ \\
BERT & Bilingual & 4 & $3 \times 10^{-5}$ & $1 \times 10^{-2}$ \\
\hline
XLM-RoBERTa & English & 8 & $1 \times 10^{-5}$ & $1 \times 10^{-5}$ \\
XLM-RoBERTa & Filipino & 8 & $3 \times 10^{-5}$ & $1 \times 10^{-2}$ \\
XLM-RoBERTa & Bilingual & 8 & $2 \times 10^{-5}$ & $1 \times 10^{-5}$ \\
\hline
RoBERTa-Tagalog & English & 8 & $3 \times 10^{-5}$ & $1 \times 10^{-2}$ \\
RoBERTa-Tagalog & Filipino & 4 & $3 \times 10^{-5}$ & $1 \times 10^{-2}$ \\
RoBERTa-Tagalog & Bilingual & 8 & $3 \times 10^{-5}$ & $1 \times 10^{-5}$ \\
\hline
NeoBERT & English & 4 & $2 \times 10^{-5}$ & $1 \times 10^{-5}$ \\
NeoBERT & Tagalog & 4 & $3 \times 10^{-5}$ & $1 \times 10^{-5}$ \\
NeoBERT & Bilingual & 4 & $3 \times 10^{-5}$ & $1 \times 10^{-5}$ \\
\hline \hline
\end{tabular}
}
\caption{Optimal hyperparameter configurations for transformers across all training regimes.}
\label{tab:best-hparams-main}
\end{table}
\subsection{Training Procedure}
Each configuration is trained independently from a fresh pretrained checkpoint to avoid parameter leakage across hyperparameter settings. Training is conducted under the following constraints: (1) a maximum of 10 epochs, which is sufficient for convergence in low-resource fine-tuning, and (2) full backpropagation over all model parameters.
\noindent
During hyperparameter search, early stopping with a patience of 3 is applied using validation macro-F1 to prevent overfitting. For final reported results, models are retrained without early stopping to ensure a consistent number of training epochs across experimental conditions.
\subsubsection{Cross-Validation Protocol}
Clinical NLP datasets are typically small and high-variance, making single train–test splits prone to unstable performance estimates. To evaluate robustness with respect to data variability, we perform stratified 10-fold cross-validation on the data for all models and training configurations. Final cross-validation performance is reported as the mean and standard deviation across folds. Importantly, cross-validation is not used for model selection in the final reported results. Instead, it is used to assess performance stability and sensitivity to data sampling. 
\subsubsection{\textbf{Evaluation Protocol and Metrics}}
For each training configuration, we evaluate under three settings. In-domain evaluation trains and tests within the same language using stratified 10-fold cross-validation. Cross-lingual zero-shot evaluation trains on each fold of one language and evaluates on the full dataset of the other language, which is never seen during training. This ensures a consistent cross-lingual test condition across folds, with variance reflecting sensitivity to training data sampling rather than target-language variability. Bilingual evaluation trains on each fold of the combined corpus ($\mathcal{D}{EN} \cup \mathcal{D}{TL}$) and evaluates on the corresponding held-out test fold. Performance is reported on the full mixed-language test fold as well as separately on English-only and Filipino-only subsets. Each configuration is trained independently.
Cross-lingual generalization is assessed by training on one language and evaluating on the other in a zero-shot setting. The combined setting evaluates models on a mixed-language test set containing both English and Filipino samples. We report Accuracy and Macro-averaged F1-score as the primary evaluation metrics. Macro-F1 is prioritized due to its equal weighting of the Healthy and Dementia classes, making it more appropriate for imbalanced clinical classification tasks where minority-class performance is critical. To quantify robustness under language shift, we define the cross-lingual generalization gap as follows
\begin{equation}
\Delta_{\text{F1}} = \left| \text{F1}_{\text{in-domain}} - \text{F1}_{\text{cross-lingual}} \right|
\end{equation}
where in-domain refers to evaluation on the same language as training (EN$\rightarrow$EN or TL$\rightarrow$TL), and cross-lingual refers to zero-shot evaluation on the unseen language (EN$\rightarrow$TL or TL$\rightarrow$EN). The combined setting is reported separately to assess performance under mixed-language conditions and is excluded from the cross-lingual generalization gap computation. In addition to overall performance, we report class-wise F1 scores and Dementia-class recall to explicitly evaluate clinical sensitivity. These metrics allow us to analyze whether performance degradation is driven by reduced detection of Dementia cases or by general classification instability.
\section{Results and Discussion}
\begin{table}[H]
\centering
\small
\setlength{\tabcolsep}{4pt}
\renewcommand{\arraystretch}{1.6}
\resizebox{\columnwidth}{!}{
\begin{tabular}{llcccc}
\toprule
& & \multicolumn{3}{c}{F1 $\pm$ std / Acc $\pm$ std} & \\
Model& Train & EN & TL & Comb & Gap \\
\midrule
TF-IDF + LR & EN & $0.930 \pm 0.013$ / $0.930 \pm 0.013$ & $0.649 \pm 0.008$ / $0.684 \pm 0.006$ & $0.836 \pm 0.005$ / $0.839 \pm 0.004$ & 0.281 \\
TF-IDF + LR & TL & $0.640 \pm 0.005$ / $0.677 \pm 0.003$ & $0.971 \pm 0.015$ / $0.971 \pm 0.015$ & $0.825 \pm 0.003$ / $0.830 \pm 0.003$ & 0.331 \\
TF-IDF + LR & EN+TL & $0.932 \pm 0.014$ / $0.932 \pm 0.014$ & $0.976 \pm 0.010$ / $0.976 \pm 0.010$ & $0.954 \pm 0.011$ / $0.954 \pm 0.011$ & 0.044 \\
\midrule
BERT & EN & $0.952 \pm 0.014$ / $0.952 \pm 0.014$ & $0.455 \pm 0.012$ / $0.561 \pm 0.007$ & $0.744 \pm 0.008$ / $0.755 \pm 0.007$ & 0.497 \\
BERT & TL & $0.705 \pm 0.033$ / $0.728 \pm 0.027$ & $0.981 \pm 0.010$ / $0.981 \pm 0.010$ & $0.855 \pm 0.014$ / $0.858 \pm 0.013$ & 0.276 \\
BERT & EN+TL & $0.954 \pm 0.009$ / $0.954 \pm 0.009$ & $0.984 \pm 0.009$ / $0.984 \pm 0.009$ & $0.969 \pm 0.007$ / $0.969 \pm 0.007$ & 0.030 \\
\midrule
XLM-RoBERTa & EN & $0.948 \pm 0.017$ / $0.949 \pm 0.017$ & $0.936 \pm 0.018$ / $0.936 \pm 0.017$ & $0.942 \pm 0.016$ / $0.942 \pm 0.016$ & 0.013 \\
XLM-RoBERTa & TL & $0.825 \pm 0.031$ / $0.830 \pm 0.029$ & $0.986 \pm 0.010$ / $0.986 \pm 0.010$ & $0.905 \pm 0.019$ / $0.906 \pm 0.019$ & 0.161 \\
XLM-RoBERTa & EN+TL & $0.953 \pm 0.010$ / $0.954 \pm 0.010$ & $0.990 \pm 0.007$ / $0.990 \pm 0.007$ & $0.972 \pm 0.006$ / $0.972 \pm 0.006$ & 0.037 \\
\midrule
RoBERTa-Tagalog & EN & $0.951 \pm 0.014$ / $0.951 \pm 0.014$ & $0.934 \pm 0.005$ / $0.934 \pm 0.005$ & $0.942 \pm 0.015$ / $0.943 \pm 0.015$ & 0.017 \\
RoBERTa-Tagalog & TL & $0.769 \pm 0.045$ / $0.781 \pm 0.038$ & $0.987 \pm 0.006$ / $0.987 \pm 0.006$ & $0.882 \pm 0.018$ / $0.884 \pm 0.017$ & 0.218 \\
RoBERTa-Tagalog & EN+TL & $0.958 \pm 0.010$ / $0.958 \pm 0.010$ & $0.988 \pm 0.007$ / $0.988 \pm 0.007$ & $0.973 \pm 0.006$ / $0.973 \pm 0.006$ & 0.030 \\
\midrule
NeoBERT & EN & $0.952 \pm 0.013$ / $0.952 \pm 0.013$ & $0.617 \pm 0.109$ / $0.667 \pm 0.084$ & $0.802 \pm 0.045$ / $0.808 \pm 0.042$ & 0.335 \\
NeoBERT & TL & $0.757 \pm 0.060$ / $0.772 \pm 0.051$ & $0.979 \pm 0.011$ / $0.979 \pm 0.011$ & $0.878 \pm 0.031$ / $0.880 \pm 0.030$ & 0.222 \\
NeoBERT & EN+TL & $0.956 \pm 0.015$ / $0.956 \pm 0.015$ & $0.983 \pm 0.009$ / $0.983 \pm 0.009$ & $0.970 \pm 0.007$ / $0.970 \pm 0.007$ & 0.027 \\
\bottomrule
\end{tabular}
}
\caption{Performance across models and training settings. Cells report F1 $\pm$ std / Acc $\pm$ std. Cross-lingual gap: $\Delta F1 = |F1_{\text{in-domain}} - F1_{\text{cross-lingual}}|$}
\label{tab:main_results}
\end{table}
Table~\ref{tab:main_results} reports Macro-F1 and Accuracy across all models and training configurations. Four findings emerge consistently. 

\subsection{Cross-Lingual Failure Persists Even Under Controlled Parallel Data}
All models achieve strong in-domain performance but degrade substantially under language transfer. English-trained BERT achieves $F1 = 0.952$ on English but falls to 0.455 on Filipino ($\Delta = 0.497$), while Filipino-trained BERT achieves $F1 = 0.981$ on Filipino but drops to 0.705 on English ($\Delta = 0.276$). The asymmetry is informative: transfer into English is consistently easier than transfer into Filipino, even when the model is fine-tuned entirely on Filipino, suggesting that English remains a stronger prior in the representation space due to pretraining exposure. Fine-tuning on Filipino modifies model behavior but does not fully overwrite the English-centric geometry learned during pretraining. These results cannot be explained by domain shift, since the Filipino corpus was constructed through controlled manual translation of the same source transcripts. The degradation, therefore, isolates language transfer rather than demographic or collection confounds. The failure appears representational: models learn dementia-related patterns within a language but fail to align those patterns robustly across languages. The lexical baseline supports this interpretation. TF-IDF~+~LR achieves $F1 = 0.649$ on Filipino when trained on English and $F1 = 0.640$ on English when trained on Filipino, suggesting that some dementia-related surface markers survive translation at the lexical level. The gap relative to transformer-based models indicates that lexical overlap alone is insufficient for stable cross-lingual generalization.

\subsection{NeoBERT Improves Monolingual Fidelity While Increasing Transfer Instability}

Relative to BERT, NeoBERT achieves comparable monolingual performance but does not improve cross-lingual robustness. Despite matching BERT's in-domain performance (F1=0.952), English-trained NeoBERT degrades substantially on Filipino (F1=0.617) and exhibits the highest variance across any experiment ($\sigma = 0.109$), suggesting that architectural modernization forms tighter English-side decision boundaries that improve in-domain fidelity while reducing tolerance to linguistic variation. Filipino-trained NeoBERT transfers better into English than BERT (F1=0.757 vs.\ 0.705) but retains high instability in dementia recall (Table~\ref{tab:class_level_clinical}). Taken together, architectural improvements alone do not produce language-invariant clinical representations, and stronger monolingual specialization appears to trade off against transfer stability.

\subsection{Transfer Behavior of Multilingual and Language-Matched Pretraining}

XLM-RoBERTa and RoBERTa-Tagalog exhibit substantially smaller transfer gaps than English-only encoders, but through different mechanisms. English-trained XLM-RoBERTa achieves $F1 = 0.936$ on Filipino ($\Delta = 0.013$), the smallest gap observed under monolingual training, consistent with multilingual pretraining constructing a shared representation space that allows English fine-tuning to map naturally onto Filipino inputs. However, transfer is asymmetric: XLM-RoBERTa transfers substantially better from English to Filipino than the reverse ($\Delta = 0.013 \ vs.\ 0.161$), suggesting that English occupies a disproportionately larger share of the pretraining corpus, producing a representation space calibrated more precisely around English structure.
RoBERTa-Tagalog is more surprising. Despite being pretrained exclusively on Filipino, it achieves nearly identical English $\rightarrow$ Filipino transfer as XLM-RoBERTa (
$F1=0.934$, $\Delta = 0.017$). We attribute this to the linguistic composition of Filipino itself: conversational Filipino contains extensive English lexical borrowing, code-switching, and Anglicized syntactic structure, meaning a model pretrained deeply on Filipino encounters substantial embedded English structure even without explicit multilingual supervision. The reverse direction supports this interpretation, as RoBERTa-Tagalog degrades more sharply when English becomes the target ($\Delta = 0.218$), suggesting that embedded English structure is sufficient for English-to-Filipino transfer but weaker than explicit multilingual supervision for reconstructing robust English-side representations after Filipino fine-tuning.

\subsection{Bilingual Training Eliminates Most Cross-Lingual Bias}

Bilingual fine-tuning substantially reduces cross-lingual degradation across all architectures, with combined-set Macro-F1 converging to 0.969--0.973 and cross-lingual gaps shrinking to between 
0.027 and 0.037. Models that behaved very differently under monolingual transfer become nearly indistinguishable after bilingual supervision. NeoBERT, which exhibited the highest instability under English-only transfer, stabilizes immediately and performs comparably to XLM-RoBERTa and RoBERTa-Tagalog, suggesting that the primary bottleneck is not architectural capacity but representational alignment. The models already possess sufficient capacity to encode dementia-related linguistic structure across both languages; what differs is whether the optimization signal encourages those structures to occupy compatible regions of the embedding space.
Multilingual and language-matched models retain a small but consistent advantage, with XLM-RoBERTa and RoBERTa-Tagalog achieving the highest combined-set performance (0.972--0.973), suggesting that pretraining language coverage provides a more favorable initialization for bilingual adaptation. Bilingual TF-IDF~+~LR substantially reduces the transfer gap ($\Delta = 0.044$) but remains below all transformer models, confirming that surface lexical statistics capture part of the dementia signal but are insufficient to construct stable cross-lingual abstractions.

\subsection{Clinical Implications Under Language Shift}
\begin{table}[t]
\centering
\small
\setlength{\tabcolsep}{3pt}
\renewcommand{\arraystretch}{1.1}
\resizebox{\columnwidth}{!}{%
\begin{tabular}{llccccccccc}
\toprule
& & \multicolumn{3}{c}{Healthy} & \multicolumn{3}{c}{Dementia} & \multicolumn{3}{c}{Dem. Recall} \\
\cmidrule(lr){3-5} \cmidrule(lr){6-8} \cmidrule(lr){9-11}
Model & Train 
& EN & TL & Comb 
& EN & TL & Comb 
& EN & TL & Comb \\
\midrule
TF-IDF+LR & EN 
& $0.934$ & $0.990$ & $0.807$
& $0.926$ & $0.989$ & $0.853$
& $0.879$ & $0.983$ & $0.968$ \\
TF-IDF+LR & TL 
& $0.955$ & $0.971$ & $0.855$
& $0.952$ & $0.970$ & $0.795$
& $0.931$ & $0.952$ & $0.660$ \\
TF-IDF+LR & EN+TL 
& $0.972$ & $0.976$ & $0.956$
& $0.971$ & $0.974$ & $0.952$
& $0.957$ & $0.950$ & $0.922$ \\
\midrule
BERT & EN 
& $0.216$ & $0.953$ & $0.690$
& $0.695$ & $0.950$ & $0.798$
& $1.000$ & $0.931$ & $0.966$ \\
BERT & TL 
& $0.981$ & $0.786$ & $0.875$
& $0.981$ & $0.624$ & $0.836$
& $0.979$ & $0.455$ & $0.724$ \\
BERT & EN+TL 
& $0.984$ & $0.955$ & $0.969$
& $0.984$ & $0.954$ & $0.969$
& $0.982$ & $0.940$ & $0.961$ \\
\midrule
XLM-RoBERTa & EN 
& $0.935$ & $0.950$ & $0.942$
& $0.936$ & $0.947$ & $0.941$
& $0.937$ & $0.920$ & $0.928$ \\
XLM-RoBERTa & TL 
& $0.986$ & $0.855$ & $0.914$
& $0.985$ & $0.794$ & $0.896$
& $0.977$ & $0.661$ & $0.815$ \\
XLM-RoBERTa & EN+TL 
& $0.990$ & $0.955$ & $0.972$
& $0.989$ & $0.952$ & $0.971$
& $0.983$ & $0.931$ & $0.957$ \\
\midrule
RoBERTa-Tagalog & EN 
& $0.935$ & $0.952$ & $0.943$
& $0.934$ & $0.950$ & $0.942$
& $0.923$ & $0.928$ & $0.925$ \\
RoBERTa-Tagalog & TL 
& $0.987$ & $0.821$ & $0.896$
& $0.987$ & $0.717$ & $0.869$
& $0.981$ & $0.563$ & $0.772$ \\
RoBERTa-Tagalog & EN+TL 
& $0.988$ & $0.959$ & $0.973$
& $0.987$ & $0.957$ & $0.972$
& $0.986$ & $0.939$ & $0.963$ \\
\midrule
NeoBERT & EN 
& $0.504$ & $0.952$ & $0.774$
& $0.729$ & $0.951$ & $0.830$
& $0.915$ & $0.939$ & $0.927$ \\
NeoBERT & TL 
& $0.979$ & $0.816$ & $0.893$
& $0.978$ & $0.699$ & $0.863$
& $0.969$ & $0.544$ & $0.766$ \\
NeoBERT & EN+TL 
& $0.983$ & $0.956$ & $0.970$
& $0.983$ & $0.955$ & $0.969$
& $0.985$ & $0.938$ & $0.962$ \\
\bottomrule
\end{tabular}%
}
\caption{Class-level performance across models and training configurations. Cells report F1; Healthy/Dementia are class F1; Dem. Recall is AD recall.}
\label{tab:class_level_clinical}
\end{table}
Table~\ref{tab:class_level_clinical} shows that aggregate Macro-F1 partially obscures clinically important failure modes under language transfer. English-trained BERT maintains high dementia recall on Filipino (0.931), but Healthy-class F1 collapses to 0.216, indicating that the model predicts most Filipino samples as dementia-positive rather than transferring meaningful discriminative structure. NeoBERT exhibits a different pattern: dementia recall on Filipino averages 0.939 but with extremely large variance ($\sigma = 0.190$), reflecting inconsistent decision boundaries across fine-tuning folds rather than stable transfer. In contrast, English-trained XLM-RoBERTa and RoBERTa-Tagalog maintain balanced class F1 scores across languages with substantially lower variance, a property particularly important in screening settings where robustness across patient populations matters more than aggregate accuracy.

Bilingual training resolves most instabilities across all architectures, with all transformer models achieving dementia recall above 0.93 with consistently low variance, indicating substantially more stable decision boundaries than those learned under monolingual training.

\subsection{Comparison with Classical Baselines}
The lexical baseline shows moderate transfer in both directions: English-trained TF-IDF~+~LR achieves $F1 = 0.649$ on Filipino and 0.640 in the reverse direction, suggesting that some dementia-related surface markers are preserved across the translation boundary. Under English-only training, TF-IDF performs comparably to NeoBERT cross-lingually but remains substantially below XLM-RoBERTa and RoBERTa-Tagalog. In the bilingual setting, TF-IDF~+~LR reaches $F1 = 0.954$ while transformer models achieve 0.969--0.973, indicating that sparse lexical features are insufficient even under bilingual exposure. These results suggest that transformer gains stem not from stronger in-language modeling alone, but from learning contextual representations that transfer across languages. TF-IDF captures surface-level dementia markers but does not model the contextual interactions through which those markers manifest across languages, whereas multilingual and language-matched models better preserve these dependencies. Overall, TF-IDF retains partial transferability due to the structural persistence of dementia-related cues, but its instability relative to contextual models indicates that robust cross-lingual screening requires representation learning beyond lexical overlap.
\section{Conclusion}
Across all model families, strong in-domain performance does not translate to cross-lingual robustness, even when domain and clinical content are held constant through parallel dataset construction. This indicates that the primary barrier to multilingual clinical NLP is representational misalignment induced by language-specific pretraining rather than task complexity or domain shift. Architectural modernization alone, as shown by NeoBERT, does not yield consistent cross-lingual gains and may increase sensitivity to language shift through tighter monolingual specialization. Multilingual and language-matched pretraining provide more stable transfer, though through different mechanisms. Bilingual fine-tuning, however, largely eliminates cross-lingual performance gaps across all architectures, suggesting that linguistic coverage during task training is more influential than architectural choice in determining robustness under language shift.

These findings indicate that reliable clinical NLP in low-resource, code-switched settings depends less on model scaling or architectural modification and more on ensuring adequate linguistic coverage during training, with bilingual supervision emerging as the most effective strategy for stable and clinically consistent performance across languages.

\section{Limitations}

Several limitations of this study warrant acknowledgment. First, the Filipino dataset was constructed through controlled manual translation of English DementiaBank transcripts, as no large-scale native clinical dementia corpora exist for Filipino. Although translators preserved structural disfluencies, hesitations, and syntactic fragmentation, the translated text inherently reflects the conversational structure and semantic content of the original English source documents. The dataset's modest scale of 4,000 samples also restricts the generalizability of evaluation metrics and contributes to cross-validation variance. Acquiring larger, organically produced clinical speech from local Filipino patient cohorts in collaboration with geriatricians remains a priority for validating these findings in local clinical environments.

Second, this study focuses exclusively on text to isolate language-specific discourse patterns, setting aside acoustic and non-verbal features. In real-world clinical settings, vocal features such as pitch variance, pause duration, and phonation rate provide independent diagnostic markers of neurodegeneration not captured in a text-only framework. Integrating speech encoders such as Wav2Vec 2.0 alongside transformer-based text models represents an important future direction.

Finally, while this evaluation establishes classification baselines across a range of encoder architectures, the mechanisms driving model decisions in multilingual contexts remain opaque. Clinical deployment demands transparency, and applying local feature attribution and interpretability methods is a necessary step before these screening workflows can achieve clinical trust and real-world deployment.



\bibliography{custom}

@article{Peled-Cohen:25,
  author  = {Lotem Peled-Cohen and Roi Reichart},
  title   = {A Systematic Review of {NLP} for Dementia: Tasks, Datasets and
             Opportunities},
  journal = {Transactions of the Association for Computational Linguistics},
  volume  = {13},
  pages   = {1204--1244},
  year    = {2025},
  doi     = {10.1162/TACL.a.35},
  url     = {https://direct.mit.edu/tacl/article/doi/10.1162/TACL.a.35/133457/}
}

@inproceedings{zhang2019rmsnorm,
  title     = {Root Mean Square Layer Normalization},
  author    = {Zhang, Biao and Sennrich, Rico},
  booktitle = {Advances in Neural Information Processing Systems (NeurIPS)},
  volume    = {32},
  year      = {2019},
  doi       = {10.48550/arXiv.1910.07467}
}

@article{shazeer2020glu,
  title   = {GLU Variants Improve Transformer},
  author  = {Shazeer, Noam},
  journal = {arXiv preprint arXiv:2002.05202},
  year    = {2020},
  url     = {https://arxiv.org/abs/2002.05202}
}

@inproceedings{Villavicencio2021Twitter,
  author    = {Charlyn Nayve Villavicencio and Julio Jerison E. Macrohon and
               X. Alphonse Inbaraj and Jyh-Horng Jeng and Jyh-Cheng Hsieh},
  title     = {Twitter Sentiment Analysis towards {COVID}-19 Vaccines in the
               {Philippines} using Na\"{i}ve {Bayes}},
  journal   = {Information},
  volume    = {12},
  pages     = {204},
  year      = {2021},
  doi       = {10.3390/info12050204},
  url       = {https://www.mdpi.com/2078-2489/12/5/204}
}

@inproceedings{miranda2023tlunified,
  author    = {Miranda, Lester James V.},
  title     = {Developing a Named Entity Recognition Dataset for {Tagalog}},
  booktitle = {Proceedings of the First Workshop in South East Asian Language
               Processing},
  pages     = {13--20},
  year      = {2023},
  month     = nov,
  address   = {Nusa Dua, Bali, Indonesia},
  publisher = {Association for Computational Linguistics},
  doi       = {10.18653/v1/2023.sealp-1.2},
  url       = {https://aclanthology.org/2023.sealp-1.2/}
}

@inproceedings{cruz-cheng-2022-improving,
  title     = {Improving Large-scale Language Models and Resources for {F}ilipino},
  author    = {Cruz, Jan Christian Blaise and Cheng, Charibeth},
  booktitle = {Proceedings of the Thirteenth Language Resources and Evaluation
               Conference},
  month     = jun,
  year      = {2022},
  address   = {Marseille, France},
  publisher = {European Language Resources Association},
  url       = {https://aclanthology.org/2022.lrec-1.703},
  pages     = {6548--6555}
}

@inproceedings{batayan2025,
  author    = {Montalan, Jann Railey and Layacan, Jimson Paulo and Africa, David Demitri and Flores, Richell Isaiah S. and {Lopez II}, Michael T. and Magsajo, Theresa Denise and Cayabyab, Anjanette and Tjhi, William Chandra},
  title     = {Batayan: A {F}ilipino {NLP} Benchmark for Evaluating Large Language Models},
  booktitle = {Proceedings of the 63rd Annual Meeting of the Association for
               Computational Linguistics (Volume 1: Long Papers)},
  pages     = {31239--31273},
  year      = {2025},
  month     = jul,
  address   = {Vienna, Austria},
  publisher = {Association for Computational Linguistics},
  doi       = {10.18653/v1/2025.acl-long.1509},
  url       = {https://aclanthology.org/2025.acl-long.1509/}
}

@inproceedings{filbench2025,
  author    = {Miranda, Lester James Validad and Aco, Elyanah and Manuel, Conner G.
               and Cruz, Jan Christian Blaise and Imperial, Joseph Marvin},
  title     = {{FilBench}: Can {LLM}s Understand and Generate {Filipino}?},
  booktitle = {Proceedings of the 2025 Conference on Empirical Methods in Natural
               Language Processing},
  pages     = {2496--2529},
  year      = {2025},
  month     = nov,
  address   = {Suzhou, China},
  publisher = {Association for Computational Linguistics},
  doi       = {10.18653/v1/2025.emnlp-main.127},
  url       = {https://aclanthology.org/2025.emnlp-main.127/}
}

@InProceedings{herrera2022tweettaglish,
  author    = {Herrera, Megan and Aich, Ankit and Parde, Natalie},
  title     = {{TweetTaglish}: A Dataset for Investigating {Tagalog-English}
               Code-Switching},
  booktitle = {Proceedings of the Thirteenth Language Resources and Evaluation
               Conference},
  month     = jun,
  year      = {2022},
  address   = {Marseille, France},
  publisher = {European Language Resources Association},
  pages     = {2592--2601},
  url       = {https://aclanthology.org/2022.lrec-1.225}
}

@inproceedings{devlin2019BERT,
  author    = {Jacob Devlin and Ming-Wei Chang and Kenton Lee and Kristina Toutanova},
  title     = {{BERT}: Pre-training of Deep Bidirectional Transformers for Language
               Understanding},
  booktitle = {Proceedings of the 2019 Conference of the North American Chapter of the
               Association for Computational Linguistics: Human Language Technologies,
               Volume 1 (Long and Short Papers)},
  pages     = {4171--4186},
  year      = {2019},
  month     = jun,
  address   = {Minneapolis, Minnesota},
  publisher = {Association for Computational Linguistics},
  doi       = {10.18653/v1/N19-1423},
  url       = {https://aclanthology.org/N19-1423/}
}

@inproceedings{pappagari2020speaker,
  author    = {Raghavendra Pappagari and Jaejin Cho and Laureano Moro-Vel\'{a}zquez
               and Najim Dehak},
  title     = {Using State of the Art Speaker Recognition and Natural Language
               Processing Technologies to Detect {Alzheimer's} Disease and Assess
               its Severity},
  booktitle = {Proceedings of Interspeech 2020},
  pages     = {2177--2181},
  year      = {2020},
  doi       = {10.21437/Interspeech.2020-2587},
  url       = {https://www.isca-archive.org/interspeech_2020/pappagari20_interspeech.html}
}

@inproceedings{alsentzer2019clinical,
  author    = {Emily Alsentzer and John Murphy and William Boag and Wei-Hung Weng
               and Di Jin and Tristan Naumann and Matthew McDermott},
  title     = {Publicly Available Clinical {BERT} Embeddings},
  booktitle = {Proceedings of the 2nd Clinical Natural Language Processing Workshop},
  pages     = {72--78},
  year      = {2019},
  month     = jun,
  address   = {Minneapolis, Minnesota, USA},
  publisher = {Association for Computational Linguistics},
  doi       = {10.18653/v1/W19-1909},
  url       = {https://aclanthology.org/W19-1909/}
}

@article{mao2023adbert,
  author  = {Chengsheng Mao and Jie Xu and Luke Rasmussen and Yikuan Li and
             Prakash Adekkanattu and Jennifer Pacheco and Borna Bonakdarpour and
             Robert Vassar and Li Shen and Guoqian Jiang and Fei Wang and
             Jyotishman Pathak and Yuan Luo},
  title   = {{AD-BERT}: Using Pre-trained Language Model to Predict the Progression
             from Mild Cognitive Impairment to {Alzheimer's} Disease},
  journal = {Journal of Biomedical Informatics},
  volume  = {144},
  pages   = {104442},
  year    = {2023},
  doi     = {10.1016/j.jbi.2023.104442},
  url     = {https://www.sciencedirect.com/science/article/pii/S1532046423001636}
}

@inproceedings{pires2019multilingual,
  author    = {Telmo Pires and Eva Schlinger and Dan Garrette},
  title     = {How Multilingual is Multilingual {BERT}?},
  booktitle = {Proceedings of the 57th Annual Meeting of the Association for
               Computational Linguistics},
  pages     = {4996--5001},
  year      = {2019},
  month     = jul,
  address   = {Florence, Italy},
  publisher = {Association for Computational Linguistics},
  doi       = {10.18653/v1/P19-1493},
  url       = {https://aclanthology.org/P19-1493/}
}

@article{Fraser:2016,
  author  = {Kathleen C. Fraser and Jed A. Meltzer and Frank Rudzicz},
  title   = {Linguistic Features Identify {Alzheimer's} Disease in Narrative Speech},
  journal = {Journal of Alzheimer's Disease},
  volume  = {49},
  number  = {2},
  pages   = {407--422},
  year    = {2016},
  doi     = {10.3233/JAD-150520},
  url     = {https://pubmed.ncbi.nlm.nih.gov/26484921/}
}

@article{Segkouli:25,
  author  = {Sofia Segkouli and Mara Gkioka and Stylianos Kokkas and Konstantinos Votis
             and Sergi Valero and Andrea Miguel and Athos Antoniades and
             Emily Charalambous and George Manias},
  title   = {Linguistic Markers in Spontaneous Speech: Insights into Subjective
             Cognitive Decline (Review)},
  journal = {Healthcare},
  volume  = {13},
  number  = {22},
  pages   = {2888},
  year    = {2025},
  doi     = {10.3390/healthcare13222888},
  url     = {https://www.mdpi.com/2227-9032/13/22/2888}
}

@article{Marier:26,
  author  = {Anna Marier and Jaime Fernández Arias and Étienne Aumont and Brandon J. Hall and Arthur C. Macedo and Nesrine Rahmouni and Gleb Bezgin and Paolo Vitali and Pedro Rosa-Neto and Maxime Montembeault},
  title   = {Language Deficits across {PET}-Based {Braak} Stages of Tau Accumulation in {Alzheimer's} Disease},
  journal = {Alzheimer's \& Dementia},
  volume  = {22},
  number  = {3},
  pages   = {e71286},
  year    = {2026},
  doi     = {10.1002/alz.71286},
  url     = {https://alz-journals.onlinelibrary.wiley.com/doi/abs/10.1002/alz.71286}
}

@article{Shankar:25,
  author  = {Ravi Shankar and Anjali Bundele and Amartya Mukhopadhyay},
  title   = {A Systematic Review of Natural Language Processing Techniques for
             Early Detection of Cognitive Impairment},
  journal = {Mayo Clinic Proceedings: Digital Health},
  volume  = {3},
  pages   = {100205},
  year    = {2025},
  doi     = {10.1016/j.mcpdig.2025.100205},
  url     = {https://www.mcpdigitalhealth.org/article/S2949-7612(25)00012-4/fulltext}
}

@article{becker1994natural,
  author  = {James T. Becker and Fran\c{c}ois Boller and Oscar L. Lopez and
             Judith Saxton and Karen L. McGonigle},
  title   = {The Natural History of {Alzheimer's} Disease: Description of Study
             Cohort and Accuracy of Diagnosis},
  journal = {Archives of Neurology},
  volume  = {51},
  number  = {6},
  pages   = {585--594},
  year    = {1994},
  month   = jun,
  doi     = {10.1001/archneur.1994.00540180063015},
  url     = {https://pubmed.ncbi.nlm.nih.gov/8198470/}
}

@article{Berkes:22,
  author  = {M. Berkes and Ellen Bialystok},
  title   = {Bilingualism as a Contributor to Cognitive Reserve: What It Can Do
             and What It Cannot Do},
  journal = {American Journal of Alzheimer's Disease \& Other Dementias},
  volume  = {37},
  year    = {2022},
  doi     = {10.1177/15333175221091417},
  url     = {https://doi.org/10.1177/15333175221091417}
}

@article{Flores:19,
  author  = {Eden Regala Flores},
  title   = {A Study on Patterns and Functions of Tagalog-English Code Switching in Two Oral Discussions},
  journal = {International Journal of TESOL Studies},
  year    = {2019},
  url     = {https://www.tesolunion.org/journal/details/info/bMTku7MjBi}
}

@inproceedings{reimers2019sentence,
  author    = {Nils Reimers and Iryna Gurevych},
  title     = {Sentence-{BERT}: Sentence Embeddings using Siamese {BERT}-Networks},
  booktitle = {Proceedings of the 2019 Conference on Empirical Methods in Natural
               Language Processing and the 9th International Joint Conference on
               Natural Language Processing (EMNLP-IJCNLP)},
  pages     = {3982--3992},
  year      = {2019},
  month     = nov,
  address   = {Hong Kong, China},
  publisher = {Association for Computational Linguistics},
  doi       = {10.18653/v1/D19-1410},
  url       = {https://aclanthology.org/D19-1410/}
}

@article{luz2024adressm,
  author  = {Saturnino Luz and Fasih Haider and Davida Fromm and Ioulietta Lazarou
             and Ioannis Kompatsiaris and Brian MacWhinney},
  title   = {An Overview of the {ADReSS-M} Signal Processing Grand Challenge on
             Multilingual {Alzheimer's} Dementia Recognition Through Spontaneous Speech},
  journal = {IEEE Open Journal of Signal Processing},
  volume  = {5},
  pages   = {738--749},
  year    = {2024},
  month   = mar,
  doi     = {10.1109/OJSP.2024.3378595},
  url     = {https://pmc.ncbi.nlm.nih.gov/articles/PMC11218814/}
}

@inproceedings{tamm2023crosslingual,
  author    = {Bastiaan Tamm and Rik Vandenberghe and Hugo {Van Hamme}},
  title     = {Cross-Lingual Transfer Learning for {Alzheimer's} Detection From
               Spontaneous Speech},
  booktitle = {ICASSP 2023 -- 2023 IEEE International Conference on Acoustics,
               Speech and Signal Processing},
  pages     = {1--2},
  year      = {2023},
  doi       = {10.1109/ICASSP49357.2023.10096770},
  url       = {https://arxiv.org/abs/2303.03049}
}

@mastersthesis{deloach2026dementia,
  author  = {Kylar A. DeLoach},
  title   = {Dementia Detection in Low-Resource Languages: Evaluating Translation-Assisted Transfer Learning for Multilingual Clinical Assessment},
  school  = {Mississippi State University},
  year    = {2026},
  type    = {Honors Thesis},
  url     = {https://scholarsjunction.msstate.edu/honorstheses/205/}
}

@article{adhikari2021exploiting,
  author  = {Surabhi Adhikari and Surendrabikram Thapa and Usman Naseem and
             Priyanka Singh and Huan Huo and Gnana Bharathy and Mukesh Prasad},
  title   = {Exploiting Linguistic Information from {Nepali} Transcripts for Early
             Detection of {Alzheimer's} Disease using Natural Language Processing
             and Machine Learning Techniques},
  journal = {International Journal of Human-Computer Studies},
  volume  = {160},
  pages   = {102761},
  year    = {2022},
  doi     = {10.1016/j.ijhcs.2021.102761},
  url     = {https://www.sciencedirect.com/science/article/abs/pii/S1071581921001798}
}

@article{lu2025amis,
  author  = {Chingching Lu and Shih-Wei Chen},
  title   = {Improving Early Detection of Dementia in Low-Resource Languages:
  A Connected Speech Analysis Case Study in {Amis} Language},
  journal = {Alzheimer's \& Dementia},
  volume  = {20},
  number  = {Suppl 8},
  pages   = {e095345},
  year    = {2025},
  doi     = {10.1002/alz.095345},
  pmcid   = {PMC11713787},
  url     = {https://pmc.ncbi.nlm.nih.gov/articles/PMC11713787/}
}

@inproceedings{conneau2020xlmr,
  author    = {Alexis Conneau and Kartikay Khandelwal and Naman Goyal and
               Vishrav Chaudhary and Guillaume Wenzek and Francisco Guzm\'{a}n and
               Edouard Grave and Myle Ott and Luke Zettlemoyer and Veselin Stoyanov},
  title     = {Unsupervised Cross-lingual Representation Learning at Scale},
  booktitle = {Proceedings of the 58th Annual Meeting of the Association for
               Computational Linguistics},
  pages     = {8440--8451},
  year      = {2020},
  address   = {Online},
  publisher = {Association for Computational Linguistics},
  doi       = {10.18653/v1/2020.acl-main.747},
  url       = {https://aclanthology.org/2020.acl-main.747/}
}

@misc{lebreton2025neobert,
  author        = {Lola Le Breton and Quentin Fournier and Mariam El Mezouar and John X. Morris and Sarath Chandar},
  title         = {{NeoBERT}: A Next-Generation {BERT}},
  year          = {2025},
  eprint        = {2502.19587},
  archivePrefix = {arXiv},
  primaryClass  = {cs.CL},
  url           = {https://arxiv.org/abs/2502.19587}
}

@article{su2021roformer,
  author        = {Jianlin Su and Yu Lu and Shengfeng Pan and Bo Wen and Yunfeng Liu},
  title         = {{RoFormer}: Enhanced Transformer with Rotary Position Embedding},
  journal       = {Neurocomputing},
  year          = {2021},
  eprint        = {2104.09864},
  archivePrefix = {arXiv},
  primaryClass  = {cs.CL},
  doi           = {10.48550/arXiv.2104.09864},
  url           = {https://arxiv.org/abs/2104.09864}
}

@inproceedings{xiong2020layer,
  author        = {Ruibin Xiong and Yunchang Yang and Di He and Kai Zheng and
                   Shuxin Zheng and Chen Xing and Huishuai Zhang and Yanyan Lan and
                   Liwei Wang and Tie-Yan Liu},
  title         = {On Layer Normalization in the Transformer Architecture},
  booktitle     = {Proceedings of the 37th International Conference on Machine Learning
                   (ICML 2020)},
  year          = {2020},
  eprint        = {2002.04745},
  archivePrefix = {arXiv},
  primaryClass  = {cs.LG},
  url           = {https://arxiv.org/abs/2002.04745}
}

@inproceedings{Sculley:2015,
  author    = {D. Sculley and Gary Holt and Daniel Golovin and Eugene Davydov and
               Todd Phillips and Dietmar Ebner and Vinay Chaudhary and Michael Young
               and Jean-Fran\c{c}ois Crespo and Dan Dennison},
  title     = {Hidden Technical Debt in Machine Learning Systems},
  booktitle = {Advances in Neural Information Processing Systems 28 (NIPS 2015)},
  pages     = {2503--2511},
  year      = {2015},
  publisher = {Curran Associates, Inc.},
  url       = {https://papers.nips.cc/paper/5656-hidden-technical-debt-in-machine-learning-systems}
}

@article{Eyigoz:2020,
  author  = {Elif Eyigoz and Sachin Mathur and Mar Mishkind and
             Guillermo Cecchi and Melissa Naylor},
  title   = {Linguistic Markers Predict Onset of {Alzheimer's} Disease},
  journal = {EClinicalMedicine},
  volume  = {28},
  pages   = {100583},
  year    = {2020},
  doi     = {10.1016/j.eclinm.2020.100583}
}

@inproceedings{Mave:2018,
  author    = {Deepthi Mave and Suraj Maharjan and Thamar Solorio},
  title     = {Language Identification and Analysis of Code-Switched Social Media Text},
  booktitle = {Proceedings of the Third Workshop on Computational Approaches to
               Linguistic Code-Switching},
  pages     = {51--61},
  year      = {2018},
  month     = jul,
  address   = {Melbourne, Australia},
  publisher = {Association for Computational Linguistics},
  doi       = {10.18653/v1/W18-3206},
  url       = {https://aclanthology.org/W18-3206/}
}

@inproceedings{salesky2019fluent,
  title={Fluent Translations from Disfluent Speech in End-to-End Speech Translation},
  author={Salesky, Elizabeth and Sperber, Matthias and Waibel, Alex},
  booktitle={Proceedings of the 2019 Conference of the North American Chapter of the Association for Computational Linguistics: Human Language Technologies (NAACL-HLT)},
  pages={2786--2792},
  year={2019},
  url={https://aclanthology.org/N19-1285/}
}

@inproceedings{guo2020benchmarking,
  title     = {Benchmarking of Transformer-Based Pre-Trained Models on Social Media
               Text Classification Datasets},
  author    = {Guo, Yuxia and Dong, Xu and Al-Garadi, Mohammed Ali and Sarker, Abeed
               and Paris, Cecile and Moll{\'{a}}, Diego},
  booktitle = {Proceedings of the 18th Annual Workshop of the Australasian Language
               Technology Association},
  year      = {2020},
  url       = {https://aclanthology.org/2020.alta-1.10},
  pages     = {104--112}
}

@inproceedings{wang2012baselines,
  author    = {Sida Wang and Christopher Manning},
  title     = {Baselines and Bigrams: Simple, Good Sentiment and Topic Classification},
  booktitle = {Proceedings of the 50th Annual Meeting of the Association for
               Computational Linguistics (Volume 2: Short Papers)},
  pages     = {90--94},
  year      = {2012},
  month     = jul,
  address   = {Jeju Island, Korea},
  publisher = {Association for Computational Linguistics},
  url       = {https://aclanthology.org/P12-2018/}
}

@incollection{zhang2015character,
  author    = {Xiang Zhang and Junbo Zhao and Yann LeCun},
  title     = {Character-level Convolutional Networks for Text Classification},
  booktitle = {Advances in Neural Information Processing Systems 28 (NIPS 2015)},
  pages     = {649--657},
  year      = {2015},
  publisher = {Curran Associates, Inc.},
  url       = {https://papers.nips.cc/paper/5782-character-level-convolutional-networks-for-text-classification}
}

@inproceedings{loshchilov2019decoupled,
  author    = {Ilya Loshchilov and Frank Hutter},
  title     = {Decoupled Weight Decay Regularization},
  booktitle = {7th International Conference on Learning Representations (ICLR 2019)},
  year      = {2019},
  address   = {New Orleans, LA, USA},
  publisher = {OpenReview.net},
  url       = {https://openreview.net/forum?id=Bkg6RiCqY7}
}

@article{masters2018revisiting,
  author        = {Dominic Masters and Carlo Luschi},
  title         = {Revisiting Small Batch Training for Deep Neural Networks},
  journal       = {arXiv preprint arXiv:1804.07612},
  year          = {2018},
  eprint        = {1804.07612},
  archivePrefix = {arXiv},
  primaryClass  = {cs.LG},
  url           = {https://arxiv.org/abs/1804.07612}
}

\end{document}